# A review of the low-cost eye-tracking systems for 2010-2020

Rakhmatulin Ildar, PhD
South Ural State University, Department of Power Plants Networks and Systems
ildar.o2010@yandex.ru

**Abstract**
The manuscript presented an analysis of the work in the field of eye-tracking over the past ten years in the low-cost filed. We researched in detail the methods, algorithms, and developed hardware. To realization, this task we considered the commercial eye-tracking systems with hardware and software and Free software. Additionally, the manuscript considered advances in the neural network fields for eye-tracking tasks and problems which hold back the development of the low-cost eye-tracking system. special attention in the manuscript is given to recommendations for further research in the field of eye-tracking devices in the low-cost field.

## 1. Introduction

Eye-tracking (ET) is the process of determining the coordinates of the gaze, with the device used to determine the orientation of the optical axis of the eyeball in space. Earlier Eye-tracking was mainly used in studies of psychophysics or cognitive development, in the last decade, the reduction in the cost of technological equipment and advances in machine vision has allowed this technology to go far beyond the scope of disease diagnosis. Today Eye-tracking is used to support multimedia learning, help in browsing the web, and is widely used in real-time graphics systems, which is especially popular for video games. The main problem of modern Eye-tracking systems is their high price. Equipment with viewing angle accuracy of less than 0.5 ° has prices from several thousand dollars.

In recent years, more than 800 peer-reviewed manuscripts published on the topic of eye-tracking in popular libraries. In the last 5 years, thanks to the development of deep learning in the field of machine vision, happened a jump, which allowed for the creation of eye-tracking devices in the low-cost range. Therefore, we have set the task to study the latest achievements in the field of eye-tracking, which has involved modern technologies over the past ten years.

We considered only completed research in which presented the test results with a low-cost eye-tracking device. According to their characteristics, these devices should have been suitable for real applications. Search for articles by keywords was carried out in the following publishers: Elsevier, Taylor & Francis, Springer, Wiley, Informa. Keyword searches were also done on the Google search engine and scholar.google.com for the last 10 years. We tried to focus on modern research. If the results coincided, an earlier source was taken for review.

The following papers (Chennamma, 2013; Kar, 2017; John, 2018) in all their content are close to our manuscript. In these manuscripts, the emphasis is on comparing a small number of works without classification. As a result, commercial options for the implementation of IT tracking are not sufficiently presented. In these works, not enough information about using of neural networks in the tasks eye-tracking.

The field of using eye-tracking is extremely diverse and includes dozens of different directions. Today, the most popular areas for using an eye-tracking system in the next fields. Eye-tracking systems are used to determine correlation-impaired eye movements caused by disorders associated with multiple **psychological illnesses.** A general overview of the diagnosis of diseases is presented in the paper (Larrazabal, 2015). To determine the disease – alcoholism by eye-tracking Maurage (2020) presented a detailed review of the modern work. Bueno (2019) submitted the review which focused on the available data on the methods of ET in neurodegenerative conditions and their potential clinical impact for cognitive assessment. Similar research presented Hessels (2019) where ET was used as a tool to study the development of perceptual and cognitive processes. Robertson (2019) described the results of studies confirming that eye-tracking reveals subtle problems in understanding speech in children with dyslexia are presented.

In recent years ET is widely used in processes to improve the quality of learning process. Sun (2019) used ET in the tasks of self-sufficiency and the effectiveness of teaching students in a programming course in C. Molina (2014) presented an empirical analysis based on eye-tracking and subjective perception of students. Liu (2014) used ET to understand the process of reading students through a learning strategy with a conceptual display of eye-tracking.

Prospect direction for research in this field it is using IT for determining the physical condition of a person. Li (2020) with the help of ET researched the identification and classification of mental fatigue of construction equipment operators using wearable eye-tracking technologies.

In recent years, with the widespread adoption of ET technologies, the scope of their application has noticeably expanded and began to be applied in previously not typical areas for this. For example, Shokishalov (2019) used of the eye-tracking system in information security. Stull (2018) analyzed the presence and activity of instructors in the video with a lecture.

Despite the diversity of areas in which ET systems were used, for the most part, noncommercial or budgetary tools were used to implement this technology. This shows the need for an inexpensive eye tracking device, which is comparable in quality to laboratory equipment.

**2. Algorithm and methods in ET research tasks**

In the past, a huge number of different methods for eye tracking were used, but in the current day, only 1 method of eye-tracking is still relevant and widely used for research and commercial purposes – video oculography. The most common distance eye-trackers use the corneal reflection (CR) method. The eyes are exposed to direct invisible infrared (IR) light, which leads to the appearance of reflection in the cornea. The physiology of this process is described in detail in the manuscript (Hari, 2012).

The following works scientific have presented novelty in the development of various algorithms and methods for tracking gaze. Huang (2020) presented a new algorithm for detecting insignificant observations of the eyes with modules for detecting objects and retraining. The first is combined with faster R-CNN and can detect the bounding frames of facial components and the initial position of the eye. The recurrence module is used to refine the orientation of the eyes using the initial shape of the eyes. Fen (2018) proposed a new model for detecting significance with a combination of superpixel segmentation and eye-tracking data. For the first time, eye tracking data is introduced to reduce the number of superpixels and speed up calculations. Ozcelik (18) presented a new method where the color coding for a more effective training neural model. The root cause of the color-coding effect using eye movement data was researched. Kerr (2019) presented a real-time correction method, which implements analysis of the collected calibration data from the user using the nearest neighbor calibration points to calculate the predicted drift at the current user focus. Larsson (2016) presented the development of a method that reliably detects events in signals recorded using a mobile eye movement tracking device. The proposed method compensates for head movements recorded using an inertial measuring unit and uses a multimodal event detection algorithm.

**3. Software for ET tasks**

For last year's, many different ET systems and various software for implementing this technology were created. The next tables show the comparative characteristics of eye-tracking devices from companies that are flagships in the development of eye-tracking devices. Table 1 shows the compact devices.

Table 1. Compact devices for eye-tracking

|   | Name | Country | Weight, gram | Frequency, Hz | Accuracy, ° | Price, $ |
|---|------|---------|--------------|---------------|-------------|----------|
| 1 | SMI Eye Tracking Glasses | Germany | 75 | 60 | 0.5 | 40000 |
| 2 | Mobile eye tracking - Tobii Glasses | Sweden | 75 | 30 | 0.5 | 20000 |
| 3 | H6 Head Mounted Optics от ASL | USA | 339 | 60 | 0.5 | 30000 |
| 4 | Mobile Eye-XG Eye Tracking Glasses от ASL | USA | 78 | 30 | 0.5 | 20000 |
| 5 | EyeTechMobile | France | 60 | 50 | 0.5 | 30000 |
| 6 | SR Research Eyelink | Canada | 300 | 500 | 0.5 | 40000 |

In table 2 presented the stationary devices for eye-tracking that can be used only in the laboratory.

Table 2. Stationary devices for eye-tracking

|   | Name | Country | Freedom of movement, cm | Work range, cm | Frequency, Hz | Accuracy, ° | Price, $ |
|---|------|---------|--------------------------|----------------|---------------|-------------|----------|
| 1 | SMI Red | Germany | 40 x 20 | 60- 80 | 60,120 | 0,4 | 40000 |
| 2 | Tobii X-series | Sweden | 44 x 22 | 50-80 | 60,120 | 0.5 | 25000 |
| 3 | ASL D6 Remote Tracking Optics | USA | 33 x 33 см | 50.8-101.6 см | 60 | 0.5 | 40000 |
| 4 | LC Technologies EyeFollower | USA | 75 x 51 | 46-97 | 120 | 0.45° | 25000 |
| 5 | SMI Red-m | Germany | 32 x 21 | 50-75 | 60,120 | 0.5 | 35000 |
| 6 | LC Technologies EAS Binocular | USA | 40 x 20 | 43-83 | 120 | 0.4 | 15000 |
| 7 | Tobii X2-series | Sweden | 50 x 36 | 40-90 | 30,60 | 0.4 | 10000 |
| 8 | Eyetech VT2 и VT2 mini | USA | 36 x 19 | 65-100 | 80 | 0.5 | 7500 |
| 9 | Mirametrix S2 Desktop Eye Tracker | Canada | 25 x 11 | 50-80 | 60 | 0.5 | 5000 |
| 10 | Gazepoint: GP3 Desktop Eye-Tracker | Canada | 25 x 11 | 50-80 | 60 | 0.5° | 500 |

| 11 | Tobii Eye Tracker | Sweden | 40 × 30 | 50–95 | 90 | 0.5° | 100 |

The next companies engaged in the commercial production of ET: Alea Technologies Gmbh Intelligaze, EyeTech Digital Systems, H.K. EyeCan: VisionKey, HumanElektronik GmbH: SeeTech, LC Technologies: Eyegaze Edge, Opportunity Foundation of America: EagleEyes, Metrovision: VISIOBOARD, IRISBOND, PRC (Prentke Romich Company): ECOpoint, TechnoWorks CO.,LTD.: Eye communication aid

The following software are popular as image analysis: AmTech GmbH, Applied Science Laboratories, Arrington Research, Cambridge Research Systems Ltd., Chronos Vision, CLS ProFakt Ltd, easyGaze(R), EL-MAR Inc., Ergoneers Dikablis, Eye-ComEyeGuide Mobile Tracker, EyeTech Digital Systems, EyeTracking, Inc., Eyeware, Fourward Technologies, Inc., ILAB, Interactive Minds, Interactive Systems Labs, Iota AB, ISCAN, LC Technologies Inc., Mangold International, Metrovision, Mirametrix, NAC Image Technology, Ober Consulting Poland: JAZZ-novo, Ober Consulting Poland: Saccadometer, Optomotor Laboratories, Pertech Primelec, D. Florin, Seeing Machines, SensoMotoric Instruments GmbH, Skalar Medical BV, Smart Eye AB, SR Research Ltd, Synthetic Environments, Inc., TestUsability, Thomas RECORDING GmbH, Tobii Technology.

The following software are free: The Eye Tribe, Smart Eye Aurora devices, Mirametrix S2, alea technologies IG-30 Pro, EyeTech VT2, ASL series, Gazepoint GP3, Senso Motoric Instruments iViewX and Red-M System, AsTeRICS, EyeRecToo, EyeTab, EyeWriter, ITU Gaze Tracker, myEye, openEyes, Opengazer, Pupil, TrackEye, WebGazer, GAZESPEAKER, Software for the automated classification of fixations and saccades, Bink-IT, Dias Eye Tracker, Eyediya Technologies Inc., EyeSpeak by LusoVu, aEye Tracking Test, Eye Tribe Tracker, I4Control®,Magic Key, myGaze & EyeGaze Education Bundle, SentiGaze.

Of interest are open source development. CARPE , CVC Eye-Tracker, COGAIN ETU Driver, EyeMMV, iComponent, GWM: Gaze-to-Word Mapping Tool, OGAMA (OpenGazeAndMouseAnalyzer), RITCode.

As a rule, these devices are presented only with nominal technical characteristics, and it is rather difficult to judge their functionality. Only a few of the devices described above have been tested and the results published. For example, Ooms (2018), Sánchez-Ferrer (22), Luo (2019) used in research the Tobii Eye device. Dong (2016) for research used Smart Eye Pro device.

As the analysis shows that equipment with high accuracy costs tens of thousands of dollars. At the same time, there is an enormous amount of open-source source, which will allow the researcher to develop software for their technical needs.

**4. Review of low-cost eye-tracking devices**

Kassner (2014) presented the Pupil Eye Tracking System with accuracy $3^0$, which works in conjunction with standard multifunction computers: laptop, desktop or tablet was presented. This is an open platform with a starting price of € 1840. Fukuda (2011) considered the possibility of using a standard web camera from computer. In the experiments, the average horizontal error of 5 participants is 3.0 °, and the average vertical error is 1.5 ° with calibration. In appearance-based methods, gaze evaluated using machine learning (ML).

Lemahieu (2010) introduced the construction of an inexpensive eye tracking device using a fixed head setting. His method is based on the algorithm for selecting the least square ellipse from the OpenCV library, fig.1.

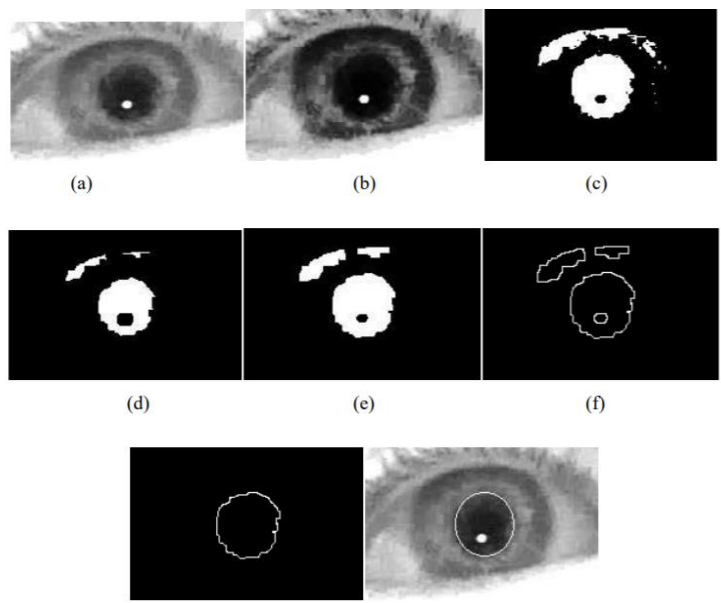

Fig. 1. Overview of the pupil detection algorithm. The original image (a) is converted with histogram alignment (b). The image is converted using a threshold function into a binary image (s). to reduce noise in the image, the extension functions (d) and erode (e) are adapted to this image. The edge detection function is calculated (f) and based on the outline of the resulting pupil shapes selected (g). The pupil is then calculated using the ellipse selection procedure (h)

The cost of the device about 100 euros with the accuracy of the device about 1.5 degrees. An experiment with text input software showed that a typing speed of 40 characters per minute.
Abdelali (2016) introduced an open-source infrastructure-Appraise that allows the use of eye-tracking. Appraise is an open-source toolkit designed to make it easier for people to evaluate machine translation. This paper mainly describes without experimental confirmations. Pavlas (2012) provided practical guidance on the creation of low-cost ET with EyeWriter and ITU GazeTracker software. The price of the device was about $ 100 with an accuracy of 2 degrees. The device is shown in fig. 2.

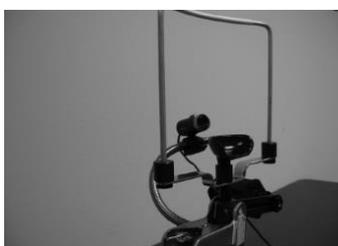

Fig.2. The prototype of the eye-tracking device

Walters (2013) presented the development of an integrated head and the eye-tracking system. The author performed Visual tracking of the head with an error of 7 mm in translation to 300 mm. and used the polar method and point comparison to determine the head position and degree of rotation.

Vargas-Cuentas (2017) developed a simple eye-tracking algorithm that does not require calibration or head holding. The system works on a portable and inexpensive tablet. The paper presented a only description of the device without characteristics and experimental studies.

Sharif (2016) in the manuscript presented the developed device - iTrace. The software of this device interacts with the eye movement tracking system and IDE to catch a glance at software artifacts and display them in their semantic meaning. In the paper presented information about experimental research.

Schneider (2011) developed an inexpensive portable eye-tracking system based on the ITU Gaze Tracker open-source software. The setup consists of a pair of built-in tracking glasses with attached cameras for recording eyes and scenes. The software has been greatly expanded and features have been added for spatial calibration, scene recording, video synchronization from the eyes and the scene, and offline tracking. The accuracy of the device is $2^0$.

Ferhat (2016) presented a device for remote tracking of gaze in visible light. The studied methods were analyzed from various points of view, such as calibration strategies, invariance of the head posture, and gaze assessment methods. The accuracy of the device is $2.35^0$. Similar research made Lee (2010), where device has accuracy $2.5^0$.

Lee (2012) studied a method for estimating the position of a three-dimensional gaze based on illuminating reflections (Purkinje images) on the surface of the cornea and lens considering the three-dimensional optical structure of the human eye model. Lin (2012) developed a new method for processing eye images in a PC-based system. With one CCD camcorder and a frame capture device that analyzes a series of images of a person's pupil when an object is looking at the screen. Lin used an automatic calibration algorithm for real-time viewing direction.

Lin (2017), introduced a new design that combines an eye-tracking device with a head gesture control module. In eye-tracking mode, the user wears glasses, and two tiny CCD cameras capture an eye image from the screen using a video capture card. In head gesture control mode, the light source projector is turned on, and the CCD camera determines the position of the light source.

**5. Review of developed eye tracking devices with neural networks**

Krafka (2016) presented software GazeCapture, which contains data from more than 1,450 people, consisting of almost 2.5 million pictures with people. Using GazeCapture, iTracker, a Krafka trained the convolutional neural network for eye tracking. The operation of the neural network is schematically shown in fig. 3.

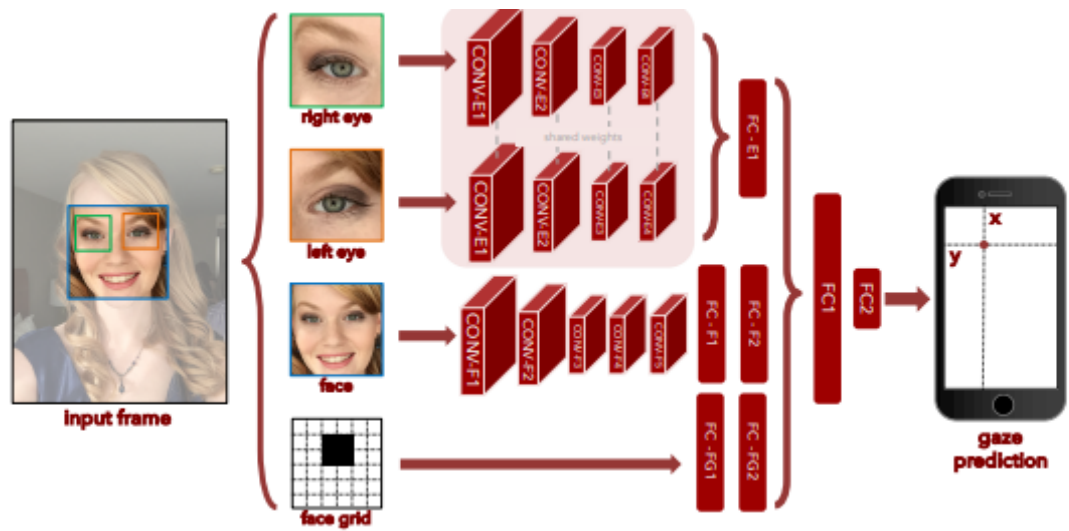

Fig. 3. The neural network diagram for GazeCapture

The author receives a model that can operate with a standard phone camera with an accuracy of $2.50^0$. Huang (2019) proposed a two-phase CNN learning strategy for combining head postures and viewing angles. The CNN architecture can reduce refit while train eye-tracking models directly with a head pose. The results of the experiment show that this method can work well when tracking the eyes with accuracy $2^0$.

**Conclusion**
A variety of eye-tracking systems, starting from hardware, and ending with software, is explained by the fact that for each application of the device there are private criteria. Despite the more than 100-year history of studying the movement behind the gaze, one of the main drawbacks of research in this area is the lack of any standards in the development of these devices. Part of the research considers the accuracy criterion of loss in degrees to be the criterion of success, where the accuracy of 0.5 degrees is considered optimal. In other papers, the most important criteria are calibration strategies, invariance of the head posture, and gaze assessment methods. In several works, comparisons are made between several eye-tracking devices.
In future studies, it is necessary to focus on standardizing research in the field of eye movement, which will allow for more competent comparisons of various research works and give a more complete picture of the situation in the field of development of these technologies.
The authors usually prefer to use standard cameras without any modification and make the novelty using various algorithms. The software in most works from open access is used.
Despite the explosive growth of interest in the subject of the use of neural networks in image processing, in the discussed papers above, as a rule, authors use standard convolution network to implement deep learning or use the tools of the OpenCV library.
The most low-cost option for developing eye tracking is the amount of 100 euros but need to keep in mind that the price of the device directly depends on the tasks and the required characteristics.

**Conflicts of Interest**: None
**Funding**: None
**Ethical Approval**: Not required